\documentclass[lettersize,journal]{IEEEtran}
\usepackage[table]{xcolor}
\usepackage{cite}   
\makeatletter
\let\NAT@parse\undefined
\makeatother

\usepackage{amsmath,amsfonts}
\usepackage{algorithmic}
\usepackage{algorithm}
\usepackage{array}
\usepackage[caption=false,font=normalsize,labelfont=sf,textfont=sf]{subfig}
\usepackage{textcomp}
\usepackage{stfloats}
\usepackage{url}
\usepackage{verbatim}
\usepackage{graphicx}
\usepackage{hyperref}
\usepackage{xcolor}
\usepackage{booktabs}  
\usepackage{bm}
\begin{document}

\title{Optimizing Local-Global Dependencies for Accurate \\3D Human Pose Estimation}


\author{Guangsheng Xu, Guoyi Zhang, Lejia Ye, Shuwei Gan, Xiaohu Zhang, and Xia Yang
	\thanks{Manuscript received xxx, xxx; revised xxx, xxx. \emph{(corresponding authors: Xiaohu Zhang, and Xia Yang.)}}
	\thanks{Guangsheng Xu, Guoyi Zhang, Lejia Ye, Shuwei Gan, Xiaohu Zhang, and Xia Yang are with the School of Aeronautics and Astronautics, Sun Yat-sen University, Shenzhen 518107, Guangdong, China. (email: xugsh6@mail2.sysu.edu.cn; zhanggy57@mail2.sysu.edu.cn; yelj8@mail2.sysu.edu.cn; ganshw3@mail2.sysu.edu.cn; zhangxiaohu@mail.sysu.edu.cn; yangxia7@mail.sysu.edu.cn)}
}

\markboth{Journal of \LaTeX\ Class Files,~Vol.~14, No.~8, August~2021}%
{Shell \MakeLowercase{\textit{et al.}}: A Sample Article Using IEEEtran.cls for IEEE Journals}



\maketitle

\begin{abstract}
Transformer-based methods have recently achieved significant success in 3D human pose estimation, owing to their strong ability to model long-range dependencies.  However, relying solely on the global attention mechanism is insufficient for capturing the fine-grained local details, which are crucial for accurate pose estimation. To address this, we propose SSR-STF, a dual-stream model that effectively integrates local features with global dependencies to enhance 3D human pose estimation. Specifically, we introduce SSRFormer, a simple yet effective module that employs the skeleton selective refine attention (SSRA) mechanism to capture fine-grained local dependencies in human pose sequences, complementing the global dependencies modeled by the Transformer. By adaptively fusing these two feature streams, SSR-STF can better learn the underlying structure of human poses, overcoming the limitations of traditional methods in local feature extraction. Extensive experiments on the Human3.6M and MPI-INF-3DHP datasets demonstrate that SSR-STF achieves state-of-the-art performance, with P1 errors of 37.4 mm and 13.2 mm respectively, outperforming existing methods in both accuracy and generalization. Furthermore, the motion representations learned by our model prove effective in downstream tasks such as human mesh recovery. Codes are available at \href{https://github.com/poker-xu/SSR-STF}{SSR-STF}.
\end{abstract}


\section{Introduction}
\label{sec:intro}

\IEEEPARstart{T}{he} 3D human pose estimation aims to predict the 3D coordinates of human joints from a single 2D image or video sequence. It plays a critical role in a wide range of applications, including human-computer interaction~\cite{svenstrup2009pose,munea2020progress}, virtual reality~\cite{mehta2017vnect}, and autonomous driving~\cite{bauer2023weakly}. The typical monocular pose estimation pipeline follows a two-stage approach, which first extracts 2D keypoints and then lifts the 2D coordinates into the 3D space. However, due to the lack of depth information and inherent ambiguity, 3D lifting methods continue to face substantial challenges in achieving robust and accurate results.\par
The Transformer architecture~\cite{vaswani2017attention} has emerged as a promising framework for monocular human pose estimation, due to its ability to capture long-range dependencies and self-attention mechanism. However, relying solely on global attention is suboptimal for pose estimation tasks because it cannot effectively capture fine-grained local dependencies, which is critical for modeling joint and motion relationships. Consequently, hybrid architectural approaches~\cite{zhao2022graformer,li2023pose, mehraban2024motionagformer} have emerged in recent years, aiming to combine local features (\emph{e.g.}, those extracted by Graph Convolution Networks, GCN) with the global dependencies captured by Transformers to enhance model performance. Although GCNs are effective at capturing interactions between vertices, they suffer from the problem of over-smoothing~\cite{chen2020measuring,liu2023skeleton}, which can undermine their ability to preserve distinct pose features. Therefore, designing networks that effectively extract local features and complement Transformers remains a key challenge that urgently needs to be addressed.\par

\begin{figure}[t]
	\centering
	\includegraphics[width=\linewidth]{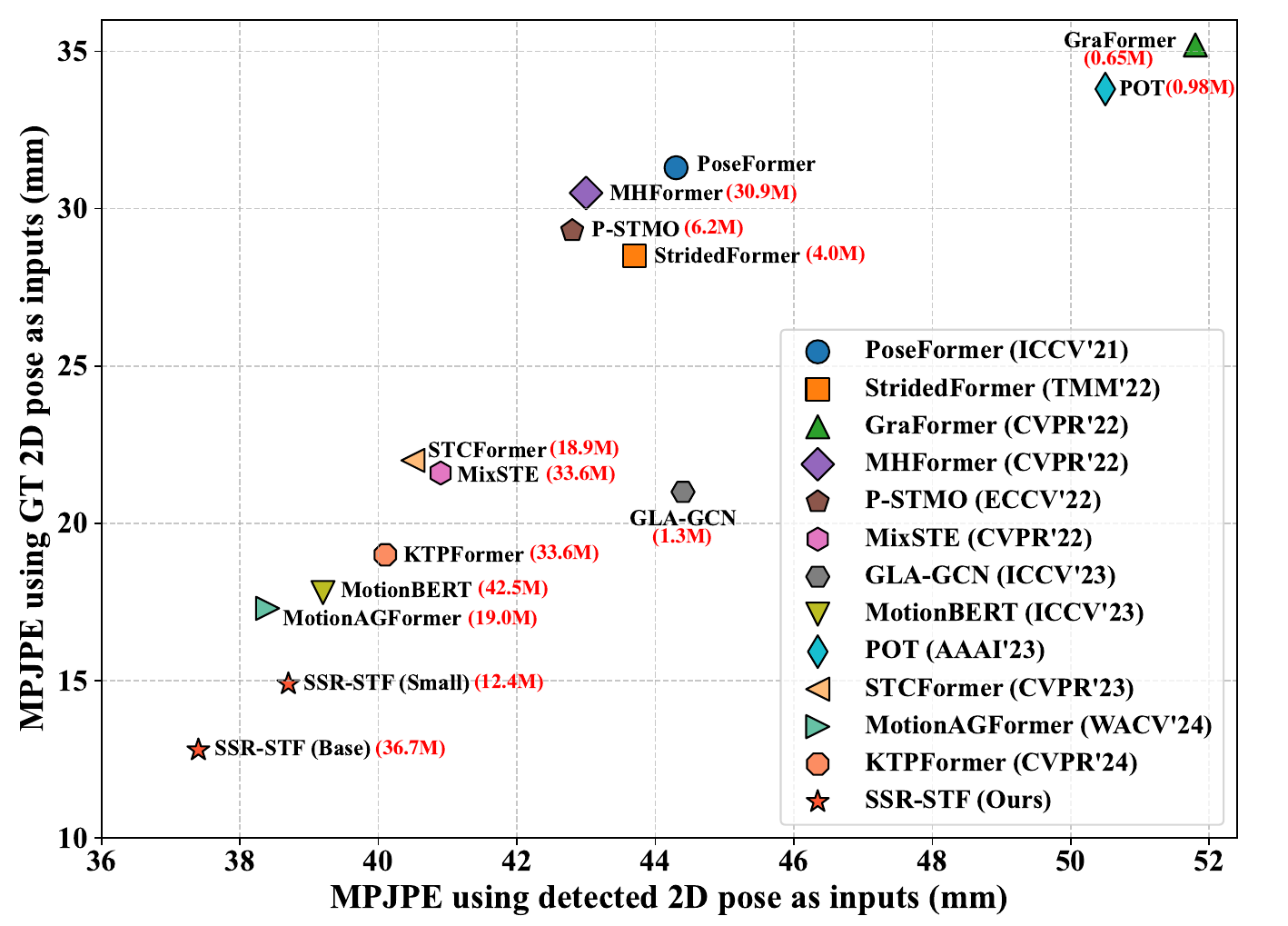}
	\caption{\textbf{MPJPE Comparison on Human3.6M dataset~\cite{ionescu2013human3} (lower is better).} 
	The horizontal and vertical axes represent MPJPE using detected 2D poses and GT 2D poses as inputs, respectively. Using GT 2D poses as input can evaluate the performance upper bound of 2D-to-3D lifting models. The number of parameters for each algorithm is also provided. Our model strikes a balance between performance and parameter efficiency, achieving a new SOTA.}  
	\label{fig:MPJPE_compare}
\end{figure}

Convolutional architectures are well-suited for capturing local details. Therefore, a potential solution is to leverage convolutional networks to extract local features from human skeleton sequences. However, in video sequences, the redundancy of temporal information is high due to the similarity of poses between adjacent frames~\cite{li2022exploiting,li2024hourglass}. The small receptive field of small-kernel convolutions makes it difficult to capture temporal motion features efficiently. More critically, the detection results in the first stage often contain errors, and the high-pass filtering nature of small-kernel convolutions may cause the model's attention to be biased toward the fluctuations induced by these detection errors. As the size of the convolutional kernel increases, the model can aggregate features over a larger receptive field, thus mitigating the negative effects of information redundancy and detection errors. However, this also leads to increased model complexity, resulting in higher parameterization and computational overhead, while making training and optimization more challenging and causing a strong tendency toward overfitting. Recent works on structural reparameterization suggest that with appropriate reparameterization, the complexity can be reduced to $\mathcal{O}(n)$~\cite{lau2024large}, or even $\mathcal{O}(log n)$~\cite{chen2024pelk}, enhancing the sparsity of operators while achieving superior performance. This approach holds great promise as a viable solution for designing networks that complement the global dependencies captured by Transformers by effectively extracting local skeletal features. \par
Monocular video 3D pose estimation requires considering feature aggregation relationships across both the temporal and spatial dimensions, where the correlations between features in each dimension are often distinct~\cite{zhang2022mixste,tang20233d,zhu2023motionbert}. Spatial correlations are primarily determined by the human skeleton, while temporal correlations depend on the redundancy in the sequential information. Standard large-kernel convolutions often fail to balance the relationship between these two dimensions effectively. To overcome this challenge, we introduce an irregular large-kernel design that assigns different weights to feature aggregation relationships in the spatial and temporal dimensions. Recent studies have highlighted that due to the static nature of convolution, its use in conjunction with Transformers often results in suboptimal performance for the Transformer model~\cite{lou2023transxnet}. To address this issue, we propose skeleton selective refine attention mechanism that breaks the inherent static nature of convolutions, enabling them to adapt to personalized inputs and form a dynamic complementary relationship with the Transformer architecture. \par
In this work, we propose SSR-STF, a dual-stream network that integrates two parallel processing streams. One stream leverages the global attention mechanism of Transformer to capture global dependencies in the skeleton sequence, while the other adopts the skeleton selective refine attention mechanism to capture local, fine-grained features. By adaptively fusing the features from both streams, SSR-STF optimizes the local-global dependencies of the pose, enabling accurate 3D human pose estimation. Extensive experiments across several public datasets demonstrate that our method achieves new state-of-the-art (SOTA) performance (See Fig.\ref{fig:MPJPE_compare}). \par
We summarize our main contributions as follows:

\begin{itemize}
	\item We propose the SSR-STF model, which effectively integrates local features captured by the SSRFormer and global dependencies captured by the Transformer, forming an efficient local-global estimator for human pose estimation.
	
	\item We introduce a novel SSRFormer module, which employs skeleton selective refine attention~(SSRA) mechanism, excelling at capturing the inherent local dependencies within human pose sequences. To the best of our knowledge, this is the first work to explore the use of large-kernel attention in skeleton-based 3D human pose estimation.
	
	\item Experimental results demonstrate the effectiveness and generalization capability of the proposed method, as it achieves state-of-the-art performance across nearly all metrics on the Human3.6M and MPI-INF-3DHP datasets.
	
	\item We further evaluate the performance of the proposed model in human mesh recovery. Experimental results demonstrate its effectiveness and substantial potential in extracting human motion representations.
\end{itemize}

\section{Related Work}
\subsection{Monocular 3D Human Pose Estimation}
Current methods for monocular 3D human pose estimation can be broadly classified into two categories: (1) Direct 3D estimation methods~\cite{moon2019camera, sun2018integral, zhou2019hemlets, wehrbein2021probabilistic, pavlakos2018ordinal} and (2) 2D-to-3D lifting methods~\cite{chen2021anatomy, liu2020attention, zheng20213d, zhang2022mixste, tang20233d,zhu2023motionbert, peng2024ktpformer}. Direct 3D estimation methods infer the joint coordinates in 3D directly from video frames without any intermediate steps. However, these methods face a trade-off between 3D pose accuracy and appearance diversity. In contrast, the 2D-to-3D lifting approach first utilizes a pre-trained 2D pose detector~\cite{chen2018cascaded, newell2016stacked, sun2019deep} to extract 2D pose estimates and then applies a separate neural network to lift these 2D poses into 3D space. 

\textbf{Transformer-based methods.}
Transformer~\cite{vaswani2017attention} has demonstrated exceptional performance in natural language processing (NLP) due to its self-attention mechanism, which effectively models long-range dependencies and captures global features. In the domain of 3D human pose estimation, PoseFormer~\cite{zheng20213d} is the first model to predict the 3D pose of the central frame entirely based on the Transformer architecture. MHFormer~\cite{li2022mhformer} proposes a method that utilizes a Transformer encoder to generate multiple pose hypotheses and synthesizes the final 3D pose by aggregating the features from these hypotheses. StridedFormer~\cite{li2022exploiting} effectively lifts the central frame by integrating 1D temporal convolutions. Unlike the above-mentioned models that only estimate the 3D pose of the central frame in a sequence, MixSTE~\cite{zhang2022mixste} alternately captures spatio-temporal features by stacking spatial and temporal transformer blocks, providing 3D estimation for each frame in the input sequence. STCFormer~\cite{tang20233d} divides the input joint features into two partitions and uses Multi-Head Self-Attention (MHSA) to capture spatial and temporal features in parallel, effectively reducing computational complexity. MotionBERT~\cite{zhu2023motionbert} utilizes a Dual-stream Spatio-temporal Transformer (DSTformer) to adaptively capture the long-range spatio-temporal relationships between skeletal joints, thereby facilitating the learning of human motion representations. D3DP~\cite{shan2023diffusion} is a diffusion-based method that recovers noisy 3D poses by combining multiple hypotheses. KTPFormer~\cite{peng2024ktpformer} introduces kinematic and trajectory priors, facilitating the effective learning of global dependencies. However, these methods pay insufficient attention to fine-grained local features.

\textbf{Hybrid methods.} 
These methods employ various modules to capture different features of the input sequences. DC-GCT~\cite{dc-gct} proposes a method that combines a Local Constraint Module based on Graph Convolutional Networks (GCN) and a Global Constraint Module based on self-attention to simultaneously leverage the local and global dependencies of the input sequence. However, this method performs poorly in feature fusion and fails to differentiate between temporal and spatial dimensions, making it less competitive compared to Transformer-based approaches. GraFormer~\cite{zhao2022graformer} stacks GCN layers with MHSA to form the GraAttention block. POT~\cite{li2023pose} incorporates the human skeleton topology into the attention mechanism in a graph-based form. However, these graph-Transformer methods~\cite{zhao2022graformer,li2023pose} focus only on the spatial correlations of the pose and neglect temporal dependencies. MotionAGFormer~\cite{mehraban2024motionagformer} introduces the Attention-GCNFormer (AGFormer) module, which combines the strengths of Transformer and GCNFormer, while considering both temporal and spatial correlations in the GCN stream, demonstrating promising performance. However, GCNs suffer from overfitting issues~\cite{chen2020measuring,liu2023skeleton} and struggle to model dependencies in the temporal dimension.

\subsection{Large Kernel Networks}
Compared to CNNs, the success of Vision Transformers (ViTs)~\cite{alexey2020image, liu2021swin} in image classification mainly stems from their ability to model long-range dependencies in the input images~\cite{luo2016understanding, ranftl2021vision, yan2021contnet, ding2022scaling}. The long-range dependencies in CNNs can be achieved through larger receptive fields and attention mechanisms~\cite{guo2023visual}. Recent studies have shown that well-designed convolutional networks with large receptive fields, such as ConvNeXt~\cite{liu2022convnet} and RepLKNet~\cite{ding2022scaling}, can strongly compete with Transformer-based models~\cite{zhang2024learningdynamiclocalcontext}. Directly increasing the size of the convolutional kernel can enlarge the effective receptive field, but it significantly increases computational cost and the number of parameters. To reduce the computational burden of large kernels, VAN~\cite{guo2023visual} proposes decomposing the large kernel into three parts: depth-wise convolution (DW), depth-wise dilation convolution (DWD), and pointwise convolution (PWConv). This allows VAN~\cite{guo2023visual} to capture longer-range dependencies than small kernels while reducing computation and parameters. The LKA proposed in VAN~\cite{guo2023visual} combines the benefits of convolution and self-attention mechanism. However, as the kernel size increases, both parameters and computation grow quadratically. To address this, LSKA~\cite{lau2024large} employs cascaded 1D depthwise convolutions (both horizontal and vertical) to limit the increase in parameters and computation, ensuring that the growth remains linear. Our focus is not on structural reparameterization, but rather on exploring large-kernel network modules that are well-suited for extracting spatio-temporal features of skeletons in the context of pose estimation tasks. 

\section{Method}
\label{method}
\begin{figure*}[t]
	\centering
	\includegraphics[width=1\linewidth]{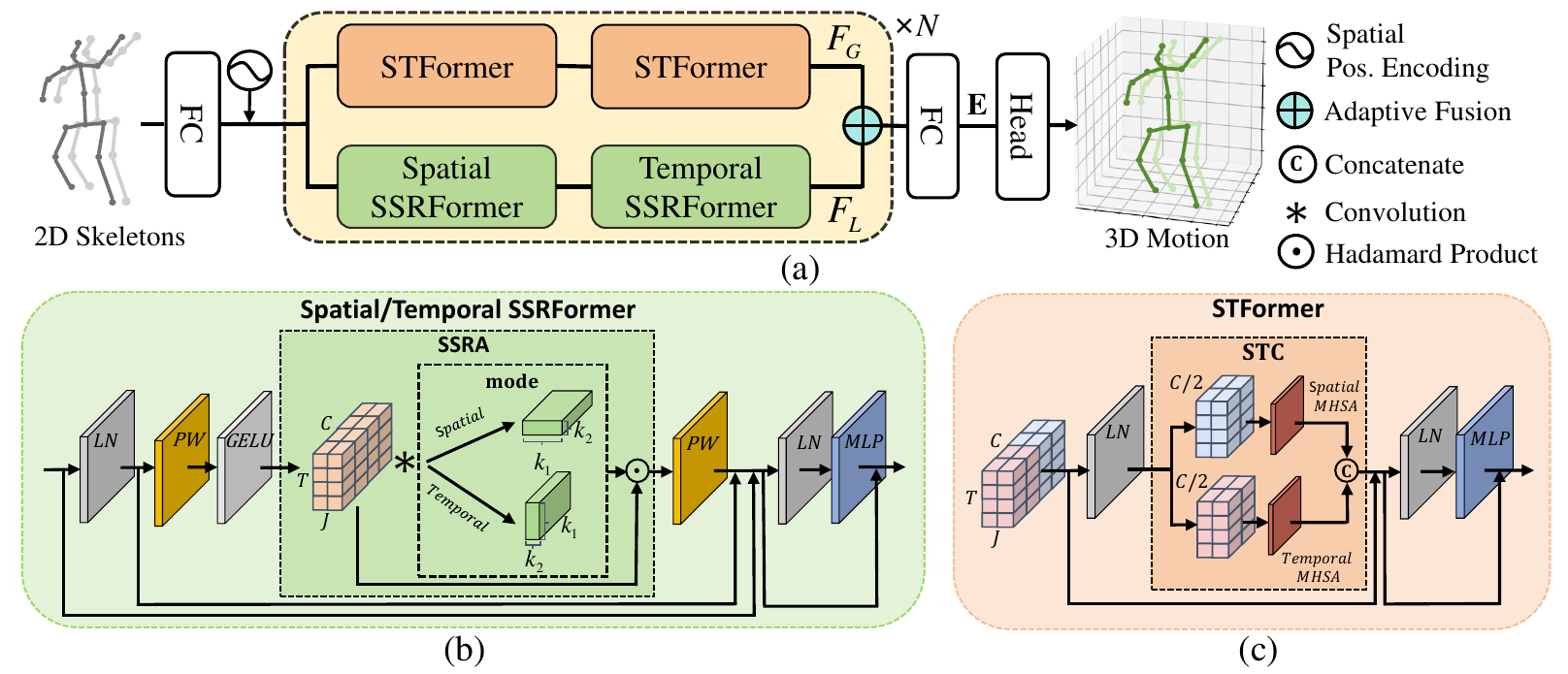}
	\caption{$(a)$ The overall architecture of SSR-STF, which is characterized by $N$ dual-stream spatio-temporal blocks, includes one stream leveraging SSRFormers and the other employing STFormers. $(b)$ Network structure of the Spatial/Temporal SSRFormer. SSRFormer employs skeleton selective refine attention mechanism~(\emph{i.e.}, SSRA) to capture the local spatio-temporal features of skeleton sequences. $(c)$ Network structure of the STFormer. STFormer adopts self-attention mechanism, excelling at capturing global dependencies.}
	\label{fig:arch}
\end{figure*}

\begin{figure}[t]
	\centering
	\includegraphics[width=\linewidth]{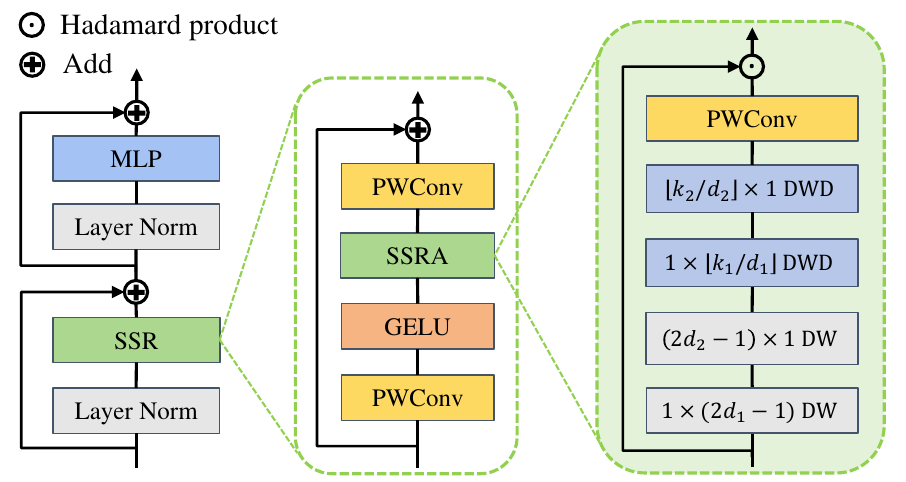}
	\caption{\textbf{SSRFormer.} 
		We employ skeleton selective refine attention mechanism to extract the spatio-temporal local features of 2D joints, as illustrated by the SSRFormer with a kernel size of $k_{1} \times k_{2}$.
	}
	\label{fig:SSRFormer}
\end{figure}

In this section, we first provide an overview of the proposed SSR-STF model for 3D human pose estimation, which features a backbone composed of SSRFormer and STFormer. We then present the detailed design of SSRFormer and the structure of STFormer. Finally, the adaptive fusion process between the local stream of SSRFormer and the global stream of STFormer is presented.

\subsection{Overall architecture}
The overall architecture of our model is illustrated in Fig.\ref{fig:arch}a.
Given a 2D sequence input with confidence scores $\mathbf{X} \in \mathbb{R} ^ {T \times J \times 3}$, where $T$ denotes the number of frames and $J$ represents the number of joints. We first project it to a high-dimensional feature $\mathbf{F}^{(0)} \in \mathbb{R} ^ {T \times J \times C}$, then a spatial positional encoding $\mathbf{P_{pos}^s} \in \mathbb{R}^{1 \times J \times C}$ is added, which helps the model differentiate between individual joints. Subsequently, we employ $N$ blocks to extract the underlying local and global information from the 2D human skeleton. Each block consists of parallel SSRFormer and STFormer modules, where the output $\mathbf{F_{L}}$ from SSRFormer and $\mathbf{F_{G}}$ from STFormer are adaptively fused to obtain $\textbf{F}^{(i)} \in \mathbb{R}^{T \times J \times C}$ ($i = 1, ..., N$). Finally, we project $\mathbf{F}^{(N)}$ into a higher-dimensional space via a linear layer followed by a $tanh$ activation, yielding the motion representation $\textbf{E} \in \mathbb{R}^{T \times J \times C_h}$. A regression head is then employed to estimate the 3D pose $\mathbf{\hat{X}} \in \mathbb{R}^{T \times J \times 3}$. 

\subsection{SSRFormer stream.}
SSRFormer (Fig.\ref{fig:arch}a) employs skeleton selective refine attention mechanism to capture local dependencies that lie between those captured by small-kernel convolutions and the Transformer, thereby serving as an effective complement to the global dependencies captured by the Transformer. SSRFormer is a MetaFormer-style~\cite{yu2022metaformer} module, with the proposed $\text{SSR}$(·) serving as the $\text{TokenMixer}$(·). The specific design of SSRFormer, which employs an equivalent \( k_1 \times k_2 \) kernel in the skeleton selective refine attention mechanism, is illustrated in Fig.\ref{fig:SSRFormer}. The output of SSRFormer, denoted as $\mathbf{Z_L}$, is given by:
\begin{equation} \label{eq:ILKA}
	\mathbf{Y_L} = \mathrm{SSR}(\mathrm{Norm}(\mathbf{X})) + \mathbf{X},
\end{equation}
\begin{equation} \label{eq:FFN_l}
	\mathbf{Z_L} = \mathrm{\sigma}(\mathrm{MLP}(\mathrm{Norm}(\mathbf{Y_L}))) + \mathbf{Y_L},
\end{equation}
Where $\sigma$(·) is the GELU~\cite{hendrycks2016gaussian} activation function, Norm(·) denotes layer normalization~\cite{lei2016layer}. 
\subsubsection{SSR Module}
The Skeleton Selective Refine (SSR) module is the core of SSRFormer, designed to enhance local information extraction. It consists of several key operations: PWConv, GELU activation, and SSRA. The mathematical formulation of the SSR module is given by:
\begin{equation}\label{eq:LSKA-atte}
	\resizebox{.9\linewidth}{!}{$\begin{array}{l}
			{\text{SSR}}(\mathbf{X})
			= \mathbf{X} + {\rm{PWConv}}({f_{{\text{SSRA}}}}({\sigma}({\rm{PWConv}}(\mathbf{X}))))
		\end{array}$}. 
\end{equation}
Where $f_{{\text{SSRA}}}$(·) denotes the skeleton selective refine attention operation. \par
\subsubsection{Large Kernel Decomposition Rule} The SSRA is built upon the Large Kernel Decomposition Rule. A $k \times k$ convolution can be decomposed into a $(2d - 1) \times (2d - 1)$ depth-wise convolution, a depth-wise dilation convolution ($(\lfloor k/d \rfloor \times \lfloor k/d \rfloor)$, with dilation rate $d$), and a $1 \times 1$ convolution~\cite{guo2023visual}. In particular, the 2D weight kernels of depth-wise convolution and depth-wise dilated convolution can be further decomposed into two cascaded 1D separable weight kernels~\cite{lau2024large}. \par
\subsubsection{Skeleton Selective Refine Attention} The feature correlations of human skeletal sequences differ across the temporal and spatial dimensions. The spatial correlations are determined by the human skeleton, while the temporal correlations depend on the redundancy of sequential information. Standard large-kernel convolutions struggle to effectively balance the relationships between these two dimensions. To address this, we propose the skeleton selective refine attention mechanism that guides the model to selectively focus on local dependencies in either the temporal or spatial dimension, thereby assigning different weights to the feature aggregation relationships in each dimension. The heart of SSR is the SSRA, as shown in Fig.\ref{fig:SSRFormer}, which selectively captures local temporal and spatial dependencies. The output of SSRA is computed as follows:
\begin{equation}\label{eq:X_bar}
	\resizebox{.7\linewidth}{!}{$\mathbf{\Bar{X}} = {\rm{DWD_2}}({\rm{DWD_1}}({\rm{DW_2}}({\rm{DW_1}}(\mathbf{X}))))$}, 
\end{equation}
\begin{equation}\label{eq:LSKAmap}
	\resizebox{.32\linewidth}{!}{$\mathbf{A} = {\rm{PWConv}}(\mathbf{\Bar{X}})$}, 
\end{equation}
\begin{equation}\label{eq:ILSKA}
	\resizebox{.35\linewidth}{!}{${f_{{\rm{SSRA}}}}(\mathbf{X}) = \mathbf{A}\odot \mathbf{X}$}.
\end{equation}
Where $\odot$ represent Hadamard product, $\rm{DW_1}$(·) and $\rm{DW_2}$(·) denote depth-wise convolution with kernel size $1 \times \left(2d_1-1\right)$ and $\left(2d_2-1\right) \times 1$ which models the local-range dependencies, respectively. $\rm{DWD_1}$(·) and $\rm{DWD_2}$(·) denote depth-wise dilated convolution with kernel size $(1 \times \lfloor k_1/d_1 \rfloor)$ and $(\lfloor k_2/d_2 \rfloor \times 1)$. Here, $\lfloor.\rfloor$ represents the floor operation, and $d_i$ is the dilation rate. $\mathbf{X}$ is the input feature map, which is reshaped upon entering and exiting SSRA(·) to selectively extract temporal or spatial information. The attention map $\mathbf{A}$ is obtained by convolving the output $\mathbf{\Bar{X}}$ of the depthwise convolution with a $1\times1$ kernel. The output of SSRA is the Hadamard product of the attention map $\mathbf{A}$ and the input feature map $\mathbf{X}$.\par
Unlike the global attention mechanism of Transformer, SSRA selectively weighs local fine-grained information, ultimately generating feature maps that enhance temporal and spatial local dependencies. SSRA and Transformer’s global attention complement each other, leveraging their respective strengths to jointly optimize the local-global dependencies in pose sequences.

\textbf{Spatial SSRFormer}. Before entering SSRA(·), the input tensor is reshaped from $\mathbb{R}^{T \times J \times C}$ to $\mathbb{R}^{C \times T \times J}$. This reordering ensures that the longer side of the irregular large kernel aligns with the \(J\)-dimension, allowing SSRFormer to better capture local dependencies in the joint space. \par
\textbf{Temporal SSRFormer}. In Temporal SSRFormer, the input tensor to SSRA(·) is reshaped to $\mathbb{R}^{C \times J \times T}$, aligning the long side of the irregular large kernel with the \( T \)-dimension, allowing SSRFormer to focus more on capturing local dependencies along the temporal dimension.

\subsection{STFormer stream.}
STFormer (Fig.\ref{fig:arch}c) leverages the attention mechanism of the Transformer to capture global spatio-temporal dependencies. Each block employs two STFormers to extract the global features $\mathbf{F_{G}}$.
For the input $\mathbf{X} \in \mathbb{R} ^ {T \times J \times C}$, the output of the STFormer stream $\mathbf{Z_G}$ can be expressed as follows:
\begin{equation} \label{eq:STC}
	\mathbf{Y_G} = \text{STC}(\mathrm{Norm}(\mathbf{X})) + \mathbf{X},
\end{equation}
\begin{equation} \label{eq:FFN_g}
	\mathbf{Z_G} = \mathrm{\sigma}(\mathrm{MLP}(\mathrm{Norm}(\mathbf{Y_G}))) + \mathbf{Y_G},
\end{equation}
Where $\text{STC}$(·)~\cite{tang20233d} refers to Spatio-Temporal Criss-cross Attention, which utilizes two parallel MHSA units to simultaneously capture long-range spatial and temporal dependencies across different channels. MHSA, a key component of the Transformer architecture, is fundamentally a mechanism that captures dependencies between different parts of the input data. The MHSA operation is defined as follows in Eq.\ref{eq:MHSA}.
\begin{equation}
	\begin{split}
		\small
		\text{MHSA}(\mathbf{Q}, \mathbf{K}, \mathbf{V})= [\text{head}_{1};...;\text{head}_{h}]\mathbf{W}^{P},  \\
		\small
		\text{head}_{i}=\text{softmax}( \frac{\mathbf{Q}^{i}(\mathbf{K}^{i})^{\prime}}{\sqrt{d_K}})\mathbf{V}^{i},
	\end{split}
	\label{eq:MHSA}
\end{equation}
where $\mathbf{W}^{P}$ is a projection parameter matrix, $h$ denotes the number of the heads, $i \in 1,\dots,h$, and $^{\prime}$ indicates matrix transpose. $\mathbf{Q}$, $\mathbf{K}$, and $\mathbf{V}$ are the query, key, and value obtained from the input tokens through linear transformations. $d_{K}$ represents the feature dimension of $\mathbf{K}$.

\subsection{Adaptive Fusion.}
In each block, STFormer leverages parallel spatial and temporal attention, i.e., STC~\cite{tang20233d}, to capture global dependencies 
$\mathbf{F_{G}}$, while SSRFormer is designed to extract local spatio-temporal features $\mathbf{F_{L}}$. Similar to MotionBERT~\cite{zhu2023motionbert}, we employ an adaptive fusion mechanism to integrate both local and global features. This is defined as follows:
\begin{equation} \label{eq:adaptivefusion}
	\resizebox{.75\linewidth}{!}{$
		\mathbf{F}^{(i)} = {\bm{\alpha_{L}}}^{(i)}\odot \mathbf{F_{L}}^{(i-1)} + {\bm{\alpha_{G}}}^{(i)}\odot \mathbf{F_{G}}^{(i-1)}
		$},
\end{equation}
where $\mathbf{F}^{(i)}$ denotes the feature embedding extracted at depth $i$, with the Hadamard product denoted by $\odot$; $\mathbf{F_{L}}^{(i-1)}$ and $\mathbf{F_{G}}^{(i-1)}$ represent the local and global features extracted from the SSRFormer and STFormer streams at depth $i-1$, respectively. The adaptive fusion weights $\bm{\alpha_{L}}$ and $\bm{\alpha_{G}}$ are defined as follows:
\begin{equation} \label{eq:adaptivefusionweights}
	\resizebox{.95\linewidth}{!}{$
		{\bm{\alpha_{L}}}^{(i)}, {\bm{\alpha_{G}}}^{(i)} = \mathrm{softmax}(\mathcal{W} \cdot \mathrm{Concat}(\mathbf{F_{L}}^{(i-1)}, \mathbf{F_{G}}^{(i-1)}))
		$},
\end{equation}
where $\mathcal{W}$ represents a learnable linear transformation.

\subsection{Loss Function.}
The joint position loss $\mathcal{L}_\text{P}$ between $\mathbf{\hat{X}}$ and GT 3D motion $\mathbf{X}$ and the velocity loss $\mathcal{L}_{\Delta \mathbf{X}}$ are given by 
\begin{equation}
	\begin{aligned}
		\small
		\label{eqn:loss_3d}
		\mathcal{L}_\text{P} = \sum\limits_{t=1}^{T} \sum\limits_{j=1}^{J} \parallel \mathbf{\hat{X}}_{t,j} - \mathbf{X}_{t,j} \parallel_2,  \\
		\mathcal{L}_{\Delta \mathbf{X}} = \sum\limits_{t=2}^{T} \sum\limits_{j=1}^{J} \parallel \Delta\mathbf{\hat{X}}_{t,j} - \Delta\mathbf{X}_{t,j} \parallel_2,
	\end{aligned}
\end{equation}
where $\Delta\mathbf{\hat{X}}_t=\mathbf{\hat{{X}}}_t - \mathbf{\hat{X}}_{t-1}$, $\Delta\mathbf{X}_t=\mathbf{X}_t - \mathbf{X}_{t-1}$. 

The total lifting loss is computed by
\begin{equation} \label{total-loss}
	\mathcal{L} = \mathcal{L}_\text{P} + \lambda_{\Delta \mathbf{X}}\mathcal{L}_{\Delta \mathbf{X}},
\end{equation}

where $\lambda_{\Delta \mathbf{X}}$ is a constant coefficient to balance position accuracy and motion smoothness.

\section{Experiment}
\label{EXPERIMENT}
We comprehensively evaluate the proposed SSR-STF on two large-scale 3D human pose estimation datasets: Human3.6M~\cite{ionescu2013human3} and MPI-INF-3DHP~\cite{mehta2017monocular}.

\begin{table*}[t]
	\caption{Quantitative comparison results with the state-of-the-art methods on Human3.6M. \textbf{Top table}: MPJPE (mm) using detected 2D pose sequences; \textbf{Bottom table}: evaluation results of P-MPJPE (mm).} 
\vspace{-5px}
\centering
\scalebox{0.8}{
	\begin{tabular}{lcc|ccccccccccccccc|c}
		\hline
		\textbf{MPJPE} &$T$ & \textbf{Publication} & Dir.          & Disc.         & Eat           & Greet         & Phone         & Photo         & Pose          & Pur.          & Sit           & SitD.         & Smoke         & Wait          & WalkD.        & Walk          & WalkT.        & Avg           \\ 
		\hline
		
		
		PoseFormer \cite{zheng20213d} &81 &ICCV'21                   & 41.5          & 44.8          & 39.8          & 42.5          & 46.5          & 51.6          & 42.1          & 42.0          & 53.3          & 60.7          & 45.5          & 43.3          & 46.1          & 31.8          & 32.2          & 44.3          \\
		StridedFormer \cite{li2022exploiting} &351 &TMM'22                  & 40.3          & 43.3          & 40.2          & 42.3          & 45.6          & 52.3          & 41.8          & 40.5          & 55.9          & 60.6          & 44.2          & 43.0          & 44.2          & 30.0          & 30.2          & 43.7          \\
		GraFormer \cite{zhao2022graformer} &1 & CVPR'22                & 45.2          & 50.8          & 48.0          & 50.0          & 54.9          & 65.0          & 48.2          & 47.1          & 60.2          & 70.0          & 51.6          & 48.7          & 54.1          & 39.7          & 43.1          & 51.8         \\
		MHFormer \cite{li2022mhformer} &351 & CVPR'22                & 39.2          & 43.1          & 40.1          & 40.9          & 44.9          & 51.2          & 40.6          & 41.3          & 53.5          & 60.3          & 43.7          & 41.1          & 43.8          & 29.8          & 30.6          & 43.0          \\
		P-STMO \cite{shan2022p} 	   &243 &ECCV'22                 & 38.9          & 42.7          & 40.4          & 41.1          & 45.6          & 49.7          & 40.9          & 39.9          & 55.5          & 59.4          & 44.9          & 42.2          & 42.7          & 29.4          & 29.4          & 42.8           \\
		MixSTE \cite{zhang2022mixste} &81 & CVPR'22                  & 39.8          & 43.0          & 38.6          & 40.1          & 43.4          & 50.6          & 40.6          & 41.4          & 52.2          & 56.7          & 43.8          & 40.8          & 43.9          & 29.4          & 30.3          & 42.4           \\
		MixSTE \cite{zhang2022mixste} &243 & CVPR'22                  & 37.6          & 40.9          & 37.3          & 39.7          & 42.3          & 49.9          & 40.1          & 39.8          & 51.7          & 55.0          & 42.1          & 39.8          & 41.0          & 27.9          & 27.9          & 40.9           \\
		DUE \cite{zhang2022uncertainty} &300 &MM’22                 & 37.9          & 41.9          & 36.8          & 39.5          & \textcolor{blue}{40.8}          & 49.2          & 40.1          & 40.7          & \textcolor{red}{47.9}          & 53.3          & \textcolor{blue}{40.2}          & 41.1          & 40.3          & 30.8          & 28.6          & 40.6          \\
		GLA-GCN \cite{yu2023gla} &243 &ICCV'23                  & 41.3          & 44.3          & 40.8          & 41.8          & 45.9          & 54.1          & 42.1          & 41.5          & 57.8          & 62.9          & 45.0          & 42.8          & 45.9          & 29.4          & 29.9          & 44.4          \\
		POT \cite{li2023pose} &1 &AAAI’23                 & 47.9           & 50.0          & 47.1           & 51.3          & 51.2          & 59.5          & 48.7          & 46.9          & 56.0          & 61.9          & 51.1          & 48.9          & 54.3          & 40.0          & 42.9          & 50.5         \\
		STCFormer \cite{tang20233d} &81  &CVPR'23                & 40.6          & 43.0          & 38.3          & 40.2          & 43.5          & 52.6          & 40.3          & 40.1          & 51.8          & 57.7         & 42.8          & 39.8          & 42.3          & 28.0          & 29.5          & 42.0          \\
		STCFormer \cite{tang20233d} &243  &CVPR'23                & 38.4          & 41.2          & 36.8          & 38.0          & 42.7          & 50.5          & 38.7          & 38.2          & 52.5          & 56.8          & 41.8          & 38.4          & 40.2          & \textcolor{blue}{26.2}          & 27.7          & 40.5          \\
		
		MotionBERT \cite{zhu2023motionbert} &243 &ICCV'23        & \textcolor{blue}{36.6}          & 39.3          & 37.8          & 33.5          & 41.4          & 49.9          & \textcolor{blue}{37.0}          & 35.5          & 50.4          & 56.5          & 41.4          & 38.2          & 37.3          & \textcolor{blue}{26.2}          & \textcolor{blue}{26.9}          & 39.2           \\
		
		D3DP* \cite{shan2023diffusion} &243 &ICCV'23            & 37.3          & 39.5          & \textcolor{blue}{35.6}          & 37.8          & 41.3          & \textcolor{blue}{48.2}          & 39.1          & 37.6          & 49.9          & \textcolor{blue}{52.8}          & 41.2          & 39.2          & 39.4          & 27.2          & 27.1          & 39.5         \\ 
		
		MotionAGFormer\cite{mehraban2024motionagformer} &243 
		&WACV'24 & 36.8 & \textcolor{blue}{38.5} & 35.9 & \textcolor{blue}{33.0} & 41.1 & 48.6 & 38.0 & \textcolor{blue}{34.8} & 49.0 & \textcolor{red}{51.4} & 40.3 & \textcolor{blue}{37.4} & \textcolor{blue}{36.3} & 27.2 & 27.2 & \textcolor{blue}{38.4} \\
		
		KTPFormer \cite{peng2024ktpformer} &243   &CVPR'24          & 37.3          & 39.2          & 35.9          & 37.6          & 42.5          & \textcolor{blue}{48.2}          & 38.6          & 39.0          & 51.4          & 55.9          & 41.6          & 39.0         & 40.0          & 27.0          & 27.4          & 40.1           \\

		\rowcolor{gray!10} Ours &27    & -      &38.8    &43.1   &40.1   &37.3    &44.3     &54.1   &40.6   &38.1    &50.9          &61.1          &44.5    & 41.7     &41.1    &30.0  &30.9   &42.4   \\ 		
		\rowcolor{gray!10} Ours &81    & -       &36.7    &39.5   &36.8   &33.6    &42.0     &50.2   &37.2   &35.6    &49.0          &58.8          &40.8    & 38.5     &37.4    &27.3  &27.7   &39.4   \\ 
		
		\rowcolor{gray!10} Ours &243   & -      & \textcolor{red}{35.4}   & \textcolor{red}{37.4}  & \textcolor{red}{35.3}   &\textcolor{red}{31.6}    & \textcolor{red}{40.6}    & \textcolor{red}{46.9}  & \textcolor{red}{36.7}  & \textcolor{red}{34.5}   & \textcolor{blue}{48.1}         & 53.9         & \textcolor{red}{39.4}   & \textcolor{red}{36.0}     & \textcolor{red}{35.1}   &\textcolor{red}{25.2}  & \textcolor{red}{25.6}  & \textcolor{red}{37.4}   \\ 
		
		\hline\hline
		\textbf{P-MPJPE} &$T$ & \textbf{Publication} & Dir.          & Disc.         & Eat           & Greet         & Phone         & Photo         & Pose          & Pur.          & Sit           & SitD.         & Smoke         & Wait          & WalkD.        & Walk          & WalkT.        & Avg            \\ 
		\hline
		
		PoseFormer \cite{zheng20213d} &81   &ICCV'21                & 34.1          & 36.1          & 34.4          & 37.2          & 36.4          & 42.2          & 34.4          & 33.6          & 45.0          & 52.5          & 37.4          & 33.8          & 37.8          & 25.6          & 27.3          & 36.5           \\
		StridedFormer \cite{li2022exploiting} &351    &TMM'22               & 32.7          & 35.5          & 32.5          & 35.4          & 35.9          & 41.6          & 33.0          & 31.9          & 45.1          & 50.1          & 36.3          & 33.5          & 35.1          & 23.9          & 25.0          & 35.2           \\
		P-STMO \cite{shan2022p} &243  &ECCV'22                & 31.3          & 35.2          & 32.9          & 33.9          & 35.4          & 39.3          & 32.5          & 31.5          & 44.6          & 48.2          & 36.3          & 32.9          & 34.4          & 23.8          & 23.9          & 34.4           \\
		MixSTE \cite{zhang2022mixste} &81 &CVPR'22                   & 32.0          & 34.2          & 31.7          & 33.7          & 34.4          & 39.2          & 32.0          & 31.8          & 42.9          & 46.9          & 35.5          & 32.0          & 34.4          & 23.6          & 25.2          & 33.9           \\
		MixSTE \cite{zhang2022mixste} &243  &CVPR'22                & 30.8          & 33.1          & \textcolor{blue}{30.3}          & 31.8          & 33.1          & 39.1          & 31.1          & 30.5          & 42.5          & \textcolor{blue}{44.5}          & 34.0          & 30.8          & 32.7          & 22.1          & 22.9          & 32.6           \\
		DUE \cite{zhang2022uncertainty} &300  &MM'22                 & 30.3          & 34.6          & \textcolor{red}{29.6}          & 31.7          & \textcolor{red}{31.6}          & 38.9          & 31.8          & 31.9          & \textcolor{red}{39.2}          & \textcolor{red}{42.8}          & \textcolor{red}{32.1}          & 32.6          & 31.4          & 25.1          & 23.8          & 32.5           \\
		GLA-GCN \cite{yu2023gla} &243 &ICCV'23                  & 32.4          & 35.3          & 32.6          & 34.2          & 35.0          & 42.1          & 32.1          & 31.9          & 45.5          & 49.5          & 36.1          & 32.4          & 35.6          & 23.5          & 24.7          & 34.8           \\
		STCFormer \cite{tang20233d} &81  &CVPR'23                & 30.4         & 33.8           & 31.1          & 31.7          & 33.5          & 39.5          & 30.8           & 30.0          & 41.8          & 45.8         & 34.3          & 30.1          & 32.8          & 21.9          & 23.4          & 32.7          \\
		STCFormer \cite{tang20233d} &243  &CVPR'23                & \textcolor{red}{29.3}        & 33.0           & 30.7          & 30.6          & 32.7          & 38.2          & \textcolor{red}{29.7}           & \textcolor{red}{28.8}         & 42.2          & 45.0         & \textcolor{blue}{33.3}          & \textcolor{red}{29.4}          & 31.5          & \textcolor{red}{20.9}          & \textcolor{red}{22.3}          & \textcolor{blue}{31.8}          \\
		
		MotionBERT \cite{zhu2023motionbert} &243 &ICCV'23        & 30.8          & 32.8          & 32.4          & 28.7          & 34.3          & 38.9          & 30.1          & 30.0          & 42.5          & 49.7          & 36.0          & 30.8          & \textcolor{red}{22.0}          & 31.7          & 23.0          & 32.9           \\
		
		MotionAGFormer\cite{mehraban2024motionagformer} &243
		&WACV'24 & 31.0 & 32.6 & 31.0 & \textcolor{blue}{27.9} & 34.0 & 38.7 & 31.5 & 30.0 & 41.4 & 45.4 & 34.8 & 30.8 & 31.3 & 22.8 & 23.2 & 32.5\\
		
		KTPFormer \cite{peng2024ktpformer} &243    &CVPR'24      & \textcolor{blue}{30.1}   & \textcolor{blue}{32.3}  & \textcolor{red}{29.6}  & 30.8   & \textcolor{blue}{32.3}    & \textcolor{red}{37.3}  & \textcolor{blue}{30.0}  & 30.2   & 41.0         & 45.3         & 33.6   & 29.9     & 31.4   & \textcolor{blue}{21.5} & \textcolor{blue}{22.6}  & 31.9  \\ 
		
		\rowcolor{gray!10} Ours &27    & -      &31.2    &34.8   &33.7   &31.2    &36.9     &42.2   &32.2   &31.4    &43.0          &50.7          &37.7    &32.7      &34.7    &24.6  &26.7   &34.9   \\ 
		\rowcolor{gray!10} Ours &81    & -      &30.4    &32.4   &31.6   &28.3    &34.7     &38.8   &30.2   &30.0    &41.3          &49.3          &35.1    &30.8      &31.8    &22.7  &24.0   &32.7   \\ 
		
		\rowcolor{gray!10} Ours &243    & -      &30.2    &\textcolor{red}{31.7}   &30.6   & \textcolor{red}{26.7}   & 33.6    & \textcolor{blue}{37.5}  &30.2   & \textcolor{blue}{29.6}   & \textcolor{blue}{40.8}         & 47.1         & 34.3   & \textcolor{blue}{29.6}     & \textcolor{blue}{30.3}   & \textcolor{blue}{21.5} & \textcolor{red}{22.3}  & \textcolor{red}{31.7}  \\ 
		
		\hline
		\multicolumn{18}{l}{\footnotesize{$T$ is the number of input frames. (*) indicates the diffusion-based methods. \textcolor{red}{Red}: Best results. \textcolor{blue}{Blue}: Runner-up results.}}
		
\end{tabular}}
\label{Tab:h36m}
\vspace{-10px}
\end{table*}

\subsection{Datasets and Evaluation Metrics}
\textbf{Human3.6M} is currently the leading benchmark for indoor 3D human pose estimation, consisting of 3.6 million video frames of 11 subjects engaged in 15 different daily activities. In accordance with the standard protocol, we utilize subjects 1, 5, 6, 7, and 8 for training, while subjects 9 and 11 are used for evaluation. To assess performance, we employ the Mean Per Joint Position Error (MPJPE) under two protocols. Protocol 1 (denoted as P1) calculates MPJPE between the estimated pose and the ground truth after aligning their root joints (hip). In contrast, Protocol 2 (denoted as P2) assesses Procrustes-MPJPE, aligning the ground truth and pose prediction through a rigid transformation.\par
\textbf{MPI-INF-3DHP} is a comprehensive dataset designed for 3D human pose estimation, capturing data in three diverse environments: green screen, non-green screen, and outdoor settings. This large-scale dataset employs evaluation metrics including MPJPE, the Percentage of Correct Keypoints (PCK) within a 150 mm threshold, and Area Under the Curve (AUC), following the methodologies established in prior research\cite{zheng20213d,shan2022p,tang20233d}. 

\subsection{Implementation Details}
Our model is implemented using the PyTorch framework and trained on two NVIDIA A100 GPUs. For the Human3.6M dataset, we use 2D pose detection results from the Stacked Hourglass~\cite{newell2016stacked}, along with the ground truth 2D poses, in accordance with~\cite{zhu2023motionbert}. For MPI-INF-3DHP, we use ground truth 2D poses, following the methodology employed in comparison baselines~\cite{tang20233d, poseformerv2, mehraban2024motionagformer}. During training, we utilize the AdamW optimizer over 90 epochs with a batch size of 12. The learning rate is initially set to 0.0006 and decayed by a factor of 0.99 for each epoch.

\subsection{Performance Comparison on Human3.6M}
We evaluate our approach against several state-of-the-art techniques on the Human3.6M dataset. To ensure a fair comparison, we include only the results from models that were not pre-trained on additional data and those where the evaluation process did not use ground truth 3D poses. The results presented in Table~\ref{Tab:h36m} show that SSR-STF achieves a P1 error of 37.4 mm and a P2 error of 31.7 mm for the estimated 2D pose. Both the P1 and P2 metrics of our method are SOTA. More importantly, our SSR-STF achieves the best performance to date in 13 out of 15 categories. \par
Table~\ref{Tab: Comparisons on human3.6 using grountruth} provides a detailed comparison between SSR-STF and SOTA models using ground-truth 2D poses as input. This setup fairly evaluates the upper bound of 2D-to-3D lifting models by eliminating the noise in 2D poses. 
SSR-STF achieves the best P1 error of 12.8 mm under $T =243$ setting, 4.5 mm ($\bf{26.0\%}$) lower than the previous SOTA model, MotionAGFormer~\cite{mehraban2024motionagformer}, validating the strong performance upper bound of the proposed model. Under $T=81$ setting, our method surpasses the second-best result by 3.4 mm ($\bf{15.3\%}$).\par
\begin{figure}[t]
\centering
\includegraphics[width=\linewidth]{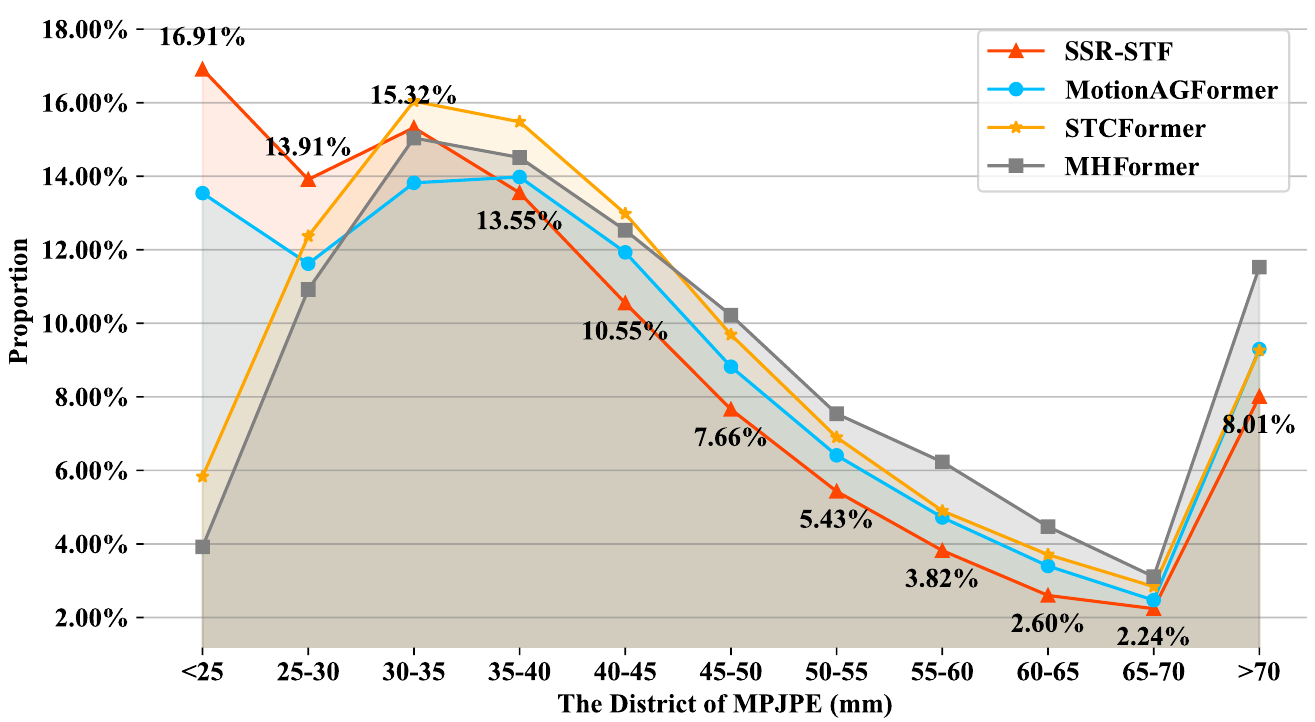}
\caption{The MPJPE distribution on Human3.6M testset, with the estimated 2D pose as input and $T=27$. The horizontal axis represents the error interval, while the vertical axis shows the proportion of poses within each error interval.} 
\label{fig:District}
\end{figure}

To evaluate the model's performance on short-time sequences, we compare the error distribution of SSR-STF with other advanced methods in Fig.\ref{fig:District}. In this experiment, all methods take the estimated 2D poses of 27 frames as input. Compared to recent transformer-based approaches, including MHFormer~\cite{li2022mhformer}, STCFormer~\cite{tang20233d}, and the hybrid method MotionAGFormer~\cite{mehraban2024motionagformer}, our method shows a significantly higher proportion of poses with MPJPE less than 30 mm and notably fewer poses with high errors. 

\begin{table*}[t]
\centering
\caption{Quantitative comparison of MPJPE (mm) with the state-of-the-art methods on Human3.6M using ground-truth (GT) 2D poses as inputs.}
\label{Tab: Comparisons on human3.6 using grountruth}
\vspace{-5px}
\scalebox{0.8}{
\begin{tabular}{lcc|ccccccccccccccc|c} 
	\hline
	\textbf{MPJPE} (GT) &$T$ & \textbf{Publication}   & Dir.          & Disc.         & Eat           & Greet         & Phone         & Photo         & Pose          & Pur.          & Sit           & SitD.         & Smoke         & Wait          & WalkD.        & Walk          & WalkT.        & Avg            \\ 
	\hline
	PoseFormer \cite{zheng20213d} &81  &ICCV'21             & 30.0          & 33.6          & 29.9          & 31.0          & 30.2          & 33.3          & 34.8          & 31.4          & 37.8          & 38.6          & 31.7          & 31.5          & 29.0          & 23.3          & 23.1          & 31.3           \\
	StridedFormer \cite{li2022exploiting} &351   &TMM'22           & 27.1          & 29.4          & 26.5          & 27.1          & 28.6          & 33.0          & 30.7          & 26.8          & 38.2          & 34.7          & 29.1          & 29.8          & 26.8          & 19.1          & 19.8          & 28.5           \\
	GraFormer \cite{zhao2022graformer} &1 & CVPR'22                & 32.0          & 38.0          & 30.4          & 34.4         & 34.7          & 43.3          & 35.2          & 31.4          & 38.0          & 46.2          & 34.2          & 35.7          & 36.1          & 27.4          & 30.6          & 35.2         \\
	MHFormer \cite{li2022mhformer} &351  & CVPR'22           & 27.7          & 32.1          & 29.1          & 28.9          & 30.0          & 33.9          & 33.0          & 31.2          & 37.0          & 39.3          & 30.0          & 31.0          & 29.4          & 22.2          & 23.0          & 30.5           \\
	P-STMO \cite{shan2022p} &243      &ECCV'22        & 28.5          & 30.1          & 28.6          & 27.9          & 29.8          & 33.2          & 31.3          & 27.8          & 36.0          & 37.4          & 29.7          & 29.5          & 28.1          & 21.0          & 21.0          & 29.3           \\
	DUE \cite{zhang2022uncertainty} &300 &MM’22                 & 22.1          & 23.1          & 20.1          & 22.7          & 21.3          & 24.1         & 23.6          & 21.6          & 26.3          & 24.8          & 21.7          & 21.4          & 21.8          & 16.7          & 18.7          & 22.0           \\
	MixSTE \cite{zhang2022mixste} &81   &CVPR'22             & 25.6          & 27.8          & 24.5          & 25.7          & 24.9          & 29.9          & 28.6          & 27.4          & 29.9          & 29.0          & 26.1          & 25.0          & 25.2          & 18.7          & 19.9          & 25.9           \\
	MixSTE \cite{zhang2022mixste} &243  &CVPR'22            & 21.6          & 22.0          & 20.4          & 21.0          & 20.8          & 24.3          & 24.7          & 21.9          & 26.9          & 24.9          & 21.2          & 21.5          & 20.8          & 14.7          & 15.7          & 21.6           \\ 
	GLA-GCN \cite{yu2023gla} &243  &ICCV'23                &  20.1         & 21.2          & 20.0          & 19.6          & 21.5         & 26.7          & 23.3          & 19.8          & 27.0          & 29.4         & 20.8          & 20.1          & 19.2          & 12.8          & 13.8          & 21.0           \\
	POT \cite{li2023pose}         &1     &AAAI’23  & 32.9            & 38.3          & 28.3           & 33.8          & 34.9          & 38.7          & 37.2          & 30.7          & 34.5          & 39.7          & 33.9          & 34.7          & 34.3          & 26.1          & 28.9          & 33.8          \\
	STCFormer \cite{tang20233d} &81  &CVPR'23                & 26.2          & 26.5          & 23.4          & 24.6          & 25.0          & 28.6          & 28.3          & 24.6          & 30.9          & 33.7         & 25.7          & 25.3          & 24.6          & 18.6          & 19.7          & 25.7           \\
	STCFormer \cite{tang20233d} &243  &CVPR'23                & 21.4          & 22.6          & 21.0          & 21.3          & 23.8          & 26.0          & 24.2          & 20.0          & 28.9          & 28.0         & 22.3          & 21.4          & 20.1          & 14.2          & 15.0          &  22.0          \\
	MotionBERT \cite{zhu2023motionbert} &243 &ICCV'23        & \textcolor{blue}{16.7}       & 19.9          &\textcolor{blue}{17.1}           & \textcolor{blue}{16.5}          & \textcolor{blue}{17.4}         & \textcolor{blue}{18.8}          & 19.3          & 20.5         & 24.0          & \textcolor{blue}{22.1}          & \textcolor{blue}{18.6}         & 16.8         & \textcolor{blue}{16.7}          & \textcolor{blue}{10.8}         & \textcolor{blue}{11.5}         & 17.8           \\
	MotionAGFormer\cite{mehraban2024motionagformer} &243 
	&WACV'24 & - &  - &  - &  - &  - &  - &  - &  - &  - & - &  - &  - &  - &  - &  - & \textcolor{blue}{17.3} \\
	
	KTPFormer \cite{peng2024ktpformer}  &81    &CVPR'24    & 22.5          & 22.4          & 21.3          & 21.4          & 21.2          & 25.5          & 24.2          & 22.4          & 24.4          & 27.5          & 22.7          & 21.4          & 21.7          & 16.3          & 17.3          & 22.2           \\
	KTPFormer \cite{peng2024ktpformer} &243    &CVPR'24    & 19.6   & \textcolor{blue}{18.6}   &18.5    & 18.1    & 18.7    & 22.1        & 20.8    & \textcolor{blue}{18.3}    & \textcolor{blue}{22.8}   & 22.4   & 18.8  & 18.1 & 18.4   & 13.9 & 15.2 & 19.0  \\
	\rowcolor{gray!10} Ours &81       & -   & 17.3          & 19.6          & 20.0          & 16.8         & 18.5          & 21.6          & \textcolor{blue}{18.8}          & 21.1          & 24.7          & 22.7          & 19.5          & \textcolor{blue}{16.6}          &  17.9         & 13.5          & 14.0          & 18.8           \\
	\rowcolor{gray!10} Ours &243      & -   & \textcolor{red}{12.4}   & \textcolor{red}{12.9}   &\textcolor{red}{14.0}    & \textcolor{red}{11.2}    & \textcolor{red}{12.9}    & \textcolor{red}{13.3}        & \textcolor{red}{13.0}    & \textcolor{red}{14.5}    & \textcolor{red}{19.4}   & \textcolor{red}{17.2}   & \textcolor{red}{14.0}  & \textcolor{red}{11.2} & \textcolor{red}{11.1}   & \textcolor{red}{7.7} & \textcolor{red}{7.7} & \textcolor{red}{12.8}  \\
	
	\hline
	\multicolumn{17}{l}{\footnotesize{$T$ is the number of input frames. \textcolor{red}{Red}: Best results. \textcolor{blue}{Blue}: Runner-up results.}}
	\end{tabular}}
	\vspace{-10px}
\end{table*}

\subsection{Performance Comparison on MPI-INF-3DHP}
To assess the generalization capability of 3D pose estimation models, we further evaluate the performance on the MPI-INF-3DHP dataset, which features more complex backgrounds. In line with previous studies~\cite{li2022mhformer, zhang2022mixste, shan2022p, tang20233d}, we use the ground-truth 2D poses as input. Table~\ref{Tab: 3DHP} presents the comparison results on the MPI-INF-3DHP test set. Our method with $T=81$ outperforms all other methods across all three metrics with PCK of 99.2\%, AUC of 88.6\%, and MPJPE of 13.2 mm. Furthermore, even with the $T=27$ setting, our method outperforms other methods using $T=81$ frames. These results highlight the robust generalization capability of our method.

\begin{table}
\caption{Performance comparisons on MPI-INF-3DHP} 
\centering
\vspace{-5px}
\scalebox{0.85}{
\begin{tabular}{lcc|ccc} 
\hline
\textbf{Method}  &T   & \textbf{Publication}            & PCK↑ & AUC↑ & MPJPE↓  \\ 
\hline
PoseFormer \cite{zheng20213d} &9  &ICCV'21     & 88.6 & 56.4 & 77.1    \\
MHFormer \cite{li2022mhformer} &9  &CVPR'22  & 93.8 & 63.3 & 58.0    \\
MixSTE \cite{zhang2022mixste} &27  &CVPR'22  & 94.4 & 66.5 & 54.9    \\
P-STMO \cite{shan2022p} &81   &ECCV'22          & 97.9 & 75.8 & 32.2    \\
PoseFormerV2 \cite{poseformerv2} &81    &CVPR'23   & 97.9 & 78.8 & 27.8    \\
GLA-GCN \cite{yu2023gla} &81    &ICCV'23       & 98.5 & 79.1 & 27.7    \\
STCFormer \cite{tang20233d} &81    &CVPR'23     & 98.7 & 83.9 & 23.1    \\ 
D3DP \cite{shan2023diffusion} &243  &ICCV'23           & 97.7 & 78.2 & 29.7    \\
MotionAGFormer \cite{peng2024ktpformer} &81 &WACV'24 & 98.2 & 85.3 & 16.2 \\
KTPFormer \cite{peng2024ktpformer} &81  &CVPR'24   & 98.9 & 85.9 & 16.7    \\

\rowcolor{gray!10} Ours  &27      & -      & \textcolor{blue}{99.1}   & \textcolor{blue}{87.5}   & \textcolor{blue}{15.0}    \\
\rowcolor{gray!10} Ours  &81      & -      & \textcolor{red}{99.2}   & \textcolor{red}{88.6}    & \textcolor{red}{13.2}    \\
\hline
\multicolumn{5}{l}{\footnotesize{Metric ↑ denotes the higher, the better, ↓ denotes the lower, the better.}}
\end{tabular}}
\label{Tab: 3DHP}
\vspace{-5px}
\end{table}

\subsection{Ablation Study}
\textbf{Architecture Hyperparameter Analysis.} We investigate the impact of different architectural hyperparameters on the performance of SSR-STF. In Table~\ref{tab:ablation_arch}, we explore the effects of varying the number of stacking modules (depth), the number of channels (width), and the value of $C_h$ on pose estimation accuracy, measured by MPJPE. The optimal configuration is highlighted in gray in the table, where our Base SSR-STF model is configured with $N=12$, $C=256$, and $C_h=512$. Additionally, we provide a Small model variant, with $N=16$, $C=128$, and $C_h=512$. The performance comparison between the two variants is shown in Fig.\ref{fig:MPJPE_compare}. Models not explicitly labeled with a variant type in this paper correspond to the Base model. \par
\textbf{Contribution of Each Component.} Table~\ref{tab:SSR-abla} provides a detailed analysis of the contribution of each component to the overall performance. The Dual stream refers to two parallel streams that utilize adaptive feature fusion. From (a) to (e) in Table~\ref{tab:SSR-abla}, we compare different combinations of components and the overall structural design. (a) and (b) are single-stream versions of STFormer and SSRFormer, respectively.  (c) is a dual-stream model where the STFormer in SSR-STF is replaced by conventional spatial and temporal Transformers. (d) represents a sequential single-stream model where SSRFormer and STFormer process data alternately. (e) represents the proposed SSR-STF, a dual-stream model where STFormer and SSRFormer operate in parallel.

As shown in Table~\ref{tab:SSR-abla}, when using only STFormer or SSRFormer, the MPJPE values are 23.1 mm and 19.9 mm, respectively. Replacing the STFormer stream in SSR-STF with a more conventional Transformer, consisting of a spatial Transformer and a temporal Transformer connected sequentially, achieves MPJPE of 15.2 mm, also surpassing the previously most competitive method. Furthermore, the sequential combination of SSRFormer and STFormer underperforms compared to the parallel fusion of the two streams. Finally, the proposed dual-stream model, SSR-STF, achieves state-of-the-art performance with MPJPE of 13.2 mm, surpassing the previous SOTA method~\cite{mehraban2024motionagformer} by 3 mm ($\bf{18.5\%}$).
 
\textbf{The Impact of Kernel Size on Performance.} Table~\ref{Tab:large kernel shapes} shows the impact of different kernel shapes in SSRA on performance. Adjusting the kernel size effectively influences SSRFormer’s ability to capture local dependencies. As shown in the table, models using small kernels (\emph{e.g.}, $3 \times 3$ convolutions) exhibit the poorest performance, while models employing skeleton selective refine attention mechanism demonstrate superior performance. On the MPI-INF-3DHP dataset, the model using the $11\times1$ kernel performs the best. However, on the Human3.6M dataset, the model with the $35 \times 11$ kernel performs slightly better, yielding a 0.3 mm lower P1 error compared to the $11 \times 1$ kernel. 
Therefore, for the experiments presented in Table~\ref{Tab:h36m}, Table~\ref{Tab: Comparisons on human3.6 using grountruth}, and Table~\ref{tab:ablation_arch} on the Human3.6M dataset, we use the equivalent irregular kernel of $35 \times 11$.

\begin{table}[t]
\caption{\textbf{Comparison of model architecture design.} All the methods are trained on Human3.6M. T is kept at 243 in all experiments.}
\begin{center}
\setlength{\tabcolsep}{3pt}
\small
\scalebox{0.9}{
\begin{tabular}{cccc|c}
\hline
Depth($N$) & Channels($C$) &$C_h$ &Param(M) & MPJPE (mm)↓  \\
\hline
6 & 256 & 512 &18.4 & 38.6 \\
8 & 256 & 512 &24.5 & 38.2 \\
10 & 256 & 512 &30.6 & 37.7 \\
\rowcolor{gray!10} 12 & 256 & 512 &36.7 & \textcolor{red}{37.4} \\
12 & 256 & 256 &36.7 & 37.4 \\
12 & 256 & 1024 &36.9 & 37.8 \\
14 & 256 & 512 &42.8 & 37.7 \\
16 & 128 & 512 &12.4 & 38.7 \\
\hline
\end{tabular}}
\end{center}
\label{tab:ablation_arch}
\end{table}

\begin{table}[h]\small
	\caption{Ablation Study Results on MPI-INF-3DHP dataset.}
	\centering
	\scalebox{0.7}{
		\begin{tabular}{l|cccc|c}
			\hline
			Strategy & STFormer & SSRFormer & Transformer & Dual stream & MPJPE↓ \\
			\hline
			(a)Baseline~\cite{tang20233d} & \checkmark & - & - & - & 23.1 \\
			(b)SSRFormer only & - & \checkmark & - & - & 19.9 \\
			(c)SSR-Transformer & - & \checkmark & \checkmark & \checkmark & 15.2 \\
			\begin{tabular}[c]{@{}c@{}}(d)SSR-STFormer \\ Sequential\end{tabular} & \checkmark & \checkmark & - & - & 14.8 \\
			\rowcolor{gray!10} (e)SSR-STF & \checkmark & \checkmark & - & \checkmark & \textcolor{red}{13.2} \\
			\hline
	\end{tabular}}
	\label{tab:SSR-abla}
\end{table}

\begin{table}
\centering
\caption{Comparison of kernel shapes in SSRA on the MPI-INF-3DHP dataset.} 
\label{Tab:large kernel shapes}
\vspace{-5px}
\scalebox{0.9}{
\begin{tabular}{cc|c} 
\hline
\textbf{Kernel shape}  & \{$k_1,d_1,k_2,d_2$\}                       &\multicolumn{1}{l}{MPJPE (mm)}  \\ 
\hline
35$\times$35   & \{$35, 3, 35, 3$\}         &14.3                                         \\
35$\times$11  & \{$35, 3, 11, 2$\}           &13.9                                        \\
23$\times$7    & \{$23, 3, 7, 2$\}         &14.2                                         \\
11$\times$11   & \{$11, 2, 11, 2$\}         &13.8                                         \\
\rowcolor{gray!10} 11$\times$1     & \{$11, 2,$ -$,$ -$\}$       &\textcolor{red}{13.2}                         \\
3$\times$3     & -       &15.9                                         \\
\hline
\end{tabular}}
\vspace{-5px}
\end{table}

\begin{figure}[t]
	\centering
	\includegraphics[width=\linewidth]{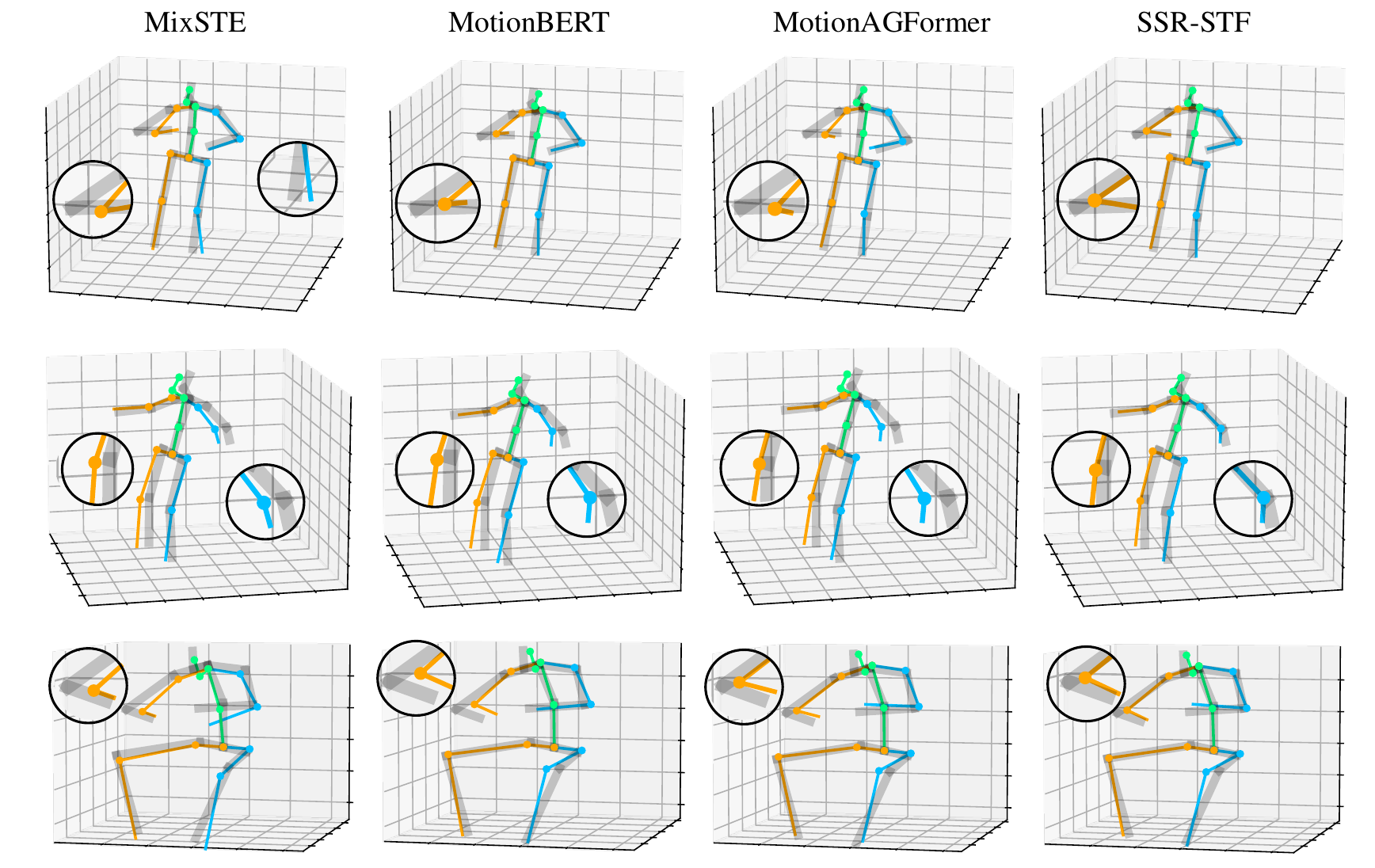}
	\caption{Qualitative comparisons of 3D pose estimation by MixSTE~\cite{zhang2022mixste},  MotionBERT~\cite{zhu2023motionbert}, MotionAGFormer~\cite{mehraban2024motionagformer} and our SSR-STF. The \textcolor{gray}{gray} skeleton is the ground-truth 3D pose. \textcolor{cyan}{Blue}, \textcolor{orange}{orange} and \textcolor{green}{green} skeletons indicate the left part, right part, and torso of the estimated body, respectively.
	}
	\label{fig:qua_compare}
\end{figure}

\subsection{Qualitative Analysis}
Fig.~\ref{fig:qua_compare} presents 3D human pose estimation results by SSR-STF and several recent Transformer-based methods, including MixSTE~\cite{zhang2022mixste}, MotionBERT~\cite{zhu2023motionbert}, and the hybrid method MotionAGFormer~\cite{mehraban2024motionagformer}. The instances are randomly selected from the evaluation set of Human3.6M. For each method, the estimated 3D human poses and ground truth coordinates are presented together for easy comparison. Overall, SSR-STF produces more accurate pose reconstructions than the other three methods across all three samples.

\subsection{Motion Representation Effectiveness in Downstream Tasks}
To validate the effectiveness of the motion representation in our model, we conducted experiments on the mesh recovery task. Specifically, we recover the human mesh by regressing the SMPL\cite{loper2023smpl} model parameters, where the 3D mesh $\mathcal{M}(\theta, \beta) \in \mathbb{R}^{6890 \times 3}$ is computed from the pose parameters $\theta \in \mathbb{R}^{72}$ and shape parameters $\beta \in \mathbb{R}^{10}$. The experimental setup follows the MotionBERT~\cite{zhu2023motionbert} framework, where the motion representation $\mathbf{E}$ is input into two separate MLP layers to regress the SMPL pose parameters $\bm{\hat{\theta}} \in \mathbb{R}^{T \times 72}$ and shape parameters $\bm{\hat{\beta}} \in \mathbb{R}^{T \times 10}$.
We trained the model from scratch on the Human3.6M~\cite{ionescu2013human3} dataset. We report MPJPE (mm) and P-MPJPE (mm) of 14 joints from $\mathbf{J} \mathcal{M}(\theta, \beta)$, where $\mathbf{J} \in \mathbb{R}^{J \times 6890}$ is a pre-defined matrix. We also report the mean per vertex error (MPVE) (mm) of the mesh $\mathcal{M}(\theta, \beta)$, which calculates the average distance between the estimated and ground truth vertices after aligning the root joint. The quantitative results are shown in Table~\ref{tab:h36m-mesh}. Our method outperforms MotionBERT~\cite{zhu2023motionbert} across three standard evaluation metrics, demonstrating a significant improvement in motion representation accuracy. This result validates the effectiveness of our proposed motion representation approach, highlighting its potential and advantage in complex tasks such as mesh recovery.

\begin{table}[]
	\caption{Quantitative comparison of human mesh recovery on Human3.6M. T denotes the clip length.}
	\centering
	\scalebox{0.85}{
		\begin{tabular}{l|ccc|ccc}
			\hline
			Method        & Input     & T &Param  & MPVE↓ & MPJPE↓ & P-MPJPE↓ \\ \hline
			MotionBERT    & 2D motion & 16 &42.5 M & 75.7     & 62.8      & 41.0         \\
			\rowcolor{gray!10}SSR-STF (ours) & 2D motion &16 &36.7 M & 68.4 & 56.3 & 35.7         \\ \hline
	\end{tabular}}
	\label{tab:h36m-mesh}
\end{table}

\section{Conclution}
\label{CONCLUSION}
We propose SSR-STF, a model that effectively integrates local features extracted by SSRFormer with global dependencies captured by the Transformer, significantly enhancing 2D-3D modeling capability. SSRFormer is a simple yet effective module that employs skeleton selective refine attention mechanism to extract local spatio-temporal features from 2D skeleton sequences, complementing the global attention mechanism of the Transformer. This forms a powerful and efficient local-global estimator for human pose estimation. Experiments on two benchmark datasets demonstrate the effectiveness and generalization ability of the proposed method, outperforming previous state-of-the-art methods on nearly all metrics. Furthermore, similar to MotionBERT~\cite{zhu2023motionbert}, our model incorporates a motion representation layer and has been evaluated on the task of human mesh recovery, demonstrating the effectiveness and potential of the motion representation in our proposed model. Future work could further explore its potential as a backbone for tasks such as video-based mesh recovery and skeleton-based action recognition.

\bibliographystyle{IEEEtran}
\bibliography{refs}

\end{document}